%% file: main.tex
\documentclass{article}

\usepackage{microtype}
\usepackage{graphicx}
\usepackage{subfigure}
\usepackage{booktabs} 

\usepackage{hyperref}

\usepackage[accepted]{icml2023}

\usepackage{amsmath}
\usepackage{amssymb}
\usepackage{mathtools}
\usepackage{amsthm}
\input{texfiles/packages}

\usepackage[capitalize,noabbrev]{cleveref}

\theoremstyle{plain}

\theoremstyle{definition}

\theoremstyle{remark}

\icmltitlerunning{Estimating the Contamination Factor's Distribution in Unsupervised Anomaly Detection}

\begin{document}

\twocolumn[
\icmltitle{Estimating the Contamination Factor's Distribution in Unsupervised Anomaly Detection}




\begin{icmlauthorlist}
\icmlauthor{Lorenzo Perini}{kul}
\icmlauthor{Paul-Christian B\"{u}rkner}{stut}
\icmlauthor{Arto Klami}{uh}
\end{icmlauthorlist}

\icmlaffiliation{kul}{DTAI lab \& Leuven.AI, Department of Computer Science, KU Leuven, Belgium}
\icmlaffiliation{stut}{Cluster of Excellence SimTech, University of Stuttgart, Germany}
\icmlaffiliation{uh}{Department of Computer Science, University of Helsinki, Finland}

\icmlcorrespondingauthor{Lorenzo Perini}{lorenzo.perini@kuleuven.be}

\icmlkeywords{Machine Learning, ICML}

\vskip 0.3in
]



\printAffiliationsAndNotice{} 

\begin{abstract}
Anomaly detection methods identify examples that do not follow the expected behaviour, typically in an unsupervised fashion, by assigning real-valued anomaly scores to the examples based on various heuristics. These scores need to be transformed into actual predictions by thresholding so that the proportion of examples marked as anomalies equals the expected proportion of anomalies, called contamination factor. Unfortunately, there are no good methods for estimating the contamination factor itself. We address this need from a Bayesian perspective, introducing a method for estimating the posterior distribution of the contamination factor for a given unlabeled dataset. We leverage several anomaly detectors to capture the basic notion of anomalousness and estimate the contamination using a specific mixture formulation. Empirically on $22$ datasets, we show that the estimated distribution is well-calibrated and that setting the threshold using the posterior mean improves the detectors' performance over several alternative methods.
\end{abstract}

\input{texfiles/introduction}
\input{texfiles/preliminaries}
\input{texfiles/methodology}
\input{texfiles/experiments}

\input{texfiles/conclusion}

\input{texfiles/acks}

\bibliography{bibliography}
\bibliographystyle{icml2023}



\end{document}

%% file: texfiles/packages.tex
\usepackage{algorithm}
\usepackage{algorithmic}
\usepackage{booktabs}       
\usepackage{amsfonts}       
\usepackage{nicefrac}       
\usepackage{microtype}      
\usepackage{xcolor}         
\usepackage{comment}
\usepackage{amsmath}
\usepackage{amsthm}
\usepackage{mathtools}

\newcommand{\ourmethod}[0]{$\gamma$\textsc{GMM}}

\usepackage{bbm}
\usepackage{footmisc}
\newcommand{\numberset}{\mathbb} %

\newcommand{\R}{\numberset{R}}
\newcommand{\E}{\numberset{E}}
\newcommand{\Pp}{\mathbb{P}}

\usepackage{cancel}

\usepackage{siunitx}
\usepackage{scalerel,stackengine} 
\stackMath

\newcommand\reallywidehat[1]{%
\savestack{\tmpbox}{\stretchto{%
  \scaleto{%
    \scalerel*[\widthof{\ensuremath{#1}}]{\kern-.6pt\bigwedge\kern-.6pt}%
    {\rule[-\textheight/2]{1ex}{\textheight}}
  }{\textheight}%
}{0.5ex}}%
\stackon[1pt]{#1}{\tmpbox}%
}
\usepackage{graphicx}
\usepackage{amsmath}
\DeclareMathOperator*{\argmax}{arg\,max}

%% file: texfiles/introduction.tex
\section{Introduction}\label{sec:introduction}

Anomaly detection aims at automatically identifying samples that do not conform to the normal behaviour, according to some notion of normality (see e.g., \citet{chandola2009anomaly}). Anomalies are often indicative of critical events such as intrusions in web networks~\citep{malaiya2018empirical}, failures in petroleum extraction~\citep{marti2015anomaly}, or breakdowns in wind and gas turbines~\citep{zaher2009online,yan2019accurate}. Such events have an associated high cost and detecting them avoids wasting time and resources. 

Typically, anomaly detection is tackled from an unsupervised perspective~\citep{maxion2000benchmarking,goldstein2016comparative,zong2018deep,perini2020quantifying,han2022adbench} because labeled samples, especially anomalies, may be expensive and difficult to acquire (e.g., you do not want to voluntarily break the equipment simply to observe anomalous behaviours), or simply rare (e.g., you may need to inspect many samples before finding an anomalous one). Unsupervised anomaly detectors exploit data-driven heuristic assumptions (e.g., anomalies are far away from normals) to assign a real-valued score to each sample denoting how anomalous it is. Using such anomaly scores enables ranking the samples from most to least anomalous.

Converting the anomaly scores into discrete predictions would practically allow the user to flag the anomalies. Commonly, one sets a decision threshold and labels samples with higher scores as anomalous and samples with lower scores as normal. However, setting the threshold is a challenging task as it cannot be tuned (e.g., by maximizing the model performance) due to the absence of labels.
One approach is to set the threshold such that the proportion of scores above it matches the dataset's \emph{contamination factor} $\gamma$, i.e. the expected proportion of anomalies. If the ranking is correct (that is, all anomalies are ranked before any normal instance) then thresholding with exactly the correct $\gamma$ correctly identifies all anomalies.
However, in most of the real-world scenarios the contamination factor is unknown.

Estimating the contamination factor $\gamma$ is challenging. 
Existing works provide an estimate by using either some normal labels~\citep{perini2020class} or domain knowledge~\citep{perini2022transferring}. Alternatively, one can directly threshold the scores through statistical threshold estimators, and derive $\gamma$ as the proportion of scores higher than the threshold. For instance, the Modified Thompson Tau test thresholder (MTT) finds the threshold through the modified Thompson Tau test~\citep{rengasamy2021towards}, while the Inter-Quartile Region thresholder (IQR) uses the third quartile plus $1.5$ times the inter-quartile region~\citep{bardet2017new}. In Section~\ref{sec:experiments} we provide a comprehensive list of estimators.

Transforming the scores into predictions using an incorrect estimate of the contamination factor (or, equivalently, an incorrect threshold) deteriorates the anomaly detector’s performance~\citep{fourure2021anomaly,emmott2015meta} and reduces the trust in the detection system. If such an estimate was coupled with a measure of uncertainty, one could take into account this uncertainty to improve decisions. Although existing methods propose Bayesian anomaly detectors~\citep{shen2010new,roberts2019bayesian,hou2022iot,heard2010bayesian}, none of them study how to transform scores into hard predictions.

Therefore, we are the first to study the estimation of the contamination factor from a Bayesian perspective. We propose \ourmethod{}, the first algorithm for estimating the contamination factor's (posterior) distribution in unlabeled anomaly detection setups. First, we use a set of unsupervised anomaly detectors to assign anomaly scores for all samples and use these scores as a new representation of the data. Second, we fit a Bayesian Gaussian Mixture model with a Dirichlet Process prior (DPGMM)~\citep{ferguson1973bayesian,rasmussen1999infinite} in this new space.
%
If we knew which components contain the anomalies, we could derive the contamination factor's posterior distribution as the distribution of the sum of such components' weights. Because we do not know this, as a third step \ourmethod{} estimates the probability that the $k$ most extreme components are jointly anomalous, and uses this information to construct the desired posterior.
The method explained in detail in Section~\ref{sec:methodology}.

In summary, we make four contributions. First, we adopt a Bayesian perspective and introduce the problem of estimating the contamination factor's posterior distribution. Second, we propose an algorithm that is able to sample from this posterior. 
Third, we demonstrate experimentally that the implied uncertainty-aware predictions are well calibrated and that taking the posterior mean as point estimate of $\gamma$ outperforms several other algorithms in common benchmarks. Finally, we show that using the posterior mean as a threshold improves the actual anomaly detection accuracy.

%% file: texfiles/preliminaries.tex
\section{Preliminaries}\label{sec:preliminaries}

Let $(\Omega, \mathcal{F}, \Pp)$ be a probability space, and $X \colon \Omega \to \R^d$ a random variable, from which a dataset $D = \{X_1,\dots,X_N\}$ of $N$ random examples is drawn. Assume that $X$ has a distribution of the form $P = (1-\gamma) \cdot P_1 + \gamma \cdot P_2$, where $P_1$ and $P_2$ are the distributions on $\R^d$ corresponding to normal examples and anomalies, respectively, and $\gamma \in [0,1]$ is the \emph{contamination factor}, i.e. the proportion of anomalies. An (unsupervised) \emph{anomaly detector} is a measurable function $f \colon \R^d \to \R$ that assigns real-valued anomaly scores $f(X)$ to the examples. Such anomaly scores follow the rule that \emph{the higher the score, the more anomalous the example}. 

A Gaussian mixture model (GMM) with $K$ components (see e.g.~\citet{roberts1998bayesian}) is a generative model defined by a distribution on a space $\R^M$ such that $p(s) = \sum_{k=1}^K \pi_k \, \mathcal{N}(s | \mu_k, \Sigma_k)$ for $s \in \R^M$, where $\mathcal{N}(s | \mu_k, \Sigma_k)$ denotes the Gaussian distribution with mean vector $\mu_k$  and covariance matrix $\Sigma_k \in \R^{M \times M}$, and $\pi_k$ are the mixing proportions such that $\sum_{k=1}^K \pi_k = 1$.
For finite mixtures, we typically have a Dirichlet prior over $\pi = [\pi_1,\dots,\pi_K]$, but Dirichlet Process (DP) priors allow treating also the number of components as unknown~\citep{gorur2010dirichlet}. 
For both cases, we need approximate inference to estimate the posterior of the model parameters.

%% file: texfiles/methodology.tex
\section{Methodology}\label{sec:methodology}
We tackle the problem: \textbf{Given} an unlabeled dataset $D$ and a set of $M$ unsupervised anomaly detectors; \textbf{Estimate} a (posterior) distribution of the contamination factor $\gamma$.

Learning from an unlabeled dataset has three key challenges. First, the absence of labels forces us to make relatively strong assumptions. Second, the anomaly detectors rely on different heuristics that may or may not hold, and their performance can hence vary significantly across datasets.
Third, we need to be careful in introducing user-specified hyperparameters, because setting them properly may be as hard as directly specifying the contamination factor.

\begin{figure*}[htbp]
\vspace{.1in}
\centerline{\includegraphics[width=1\textwidth]{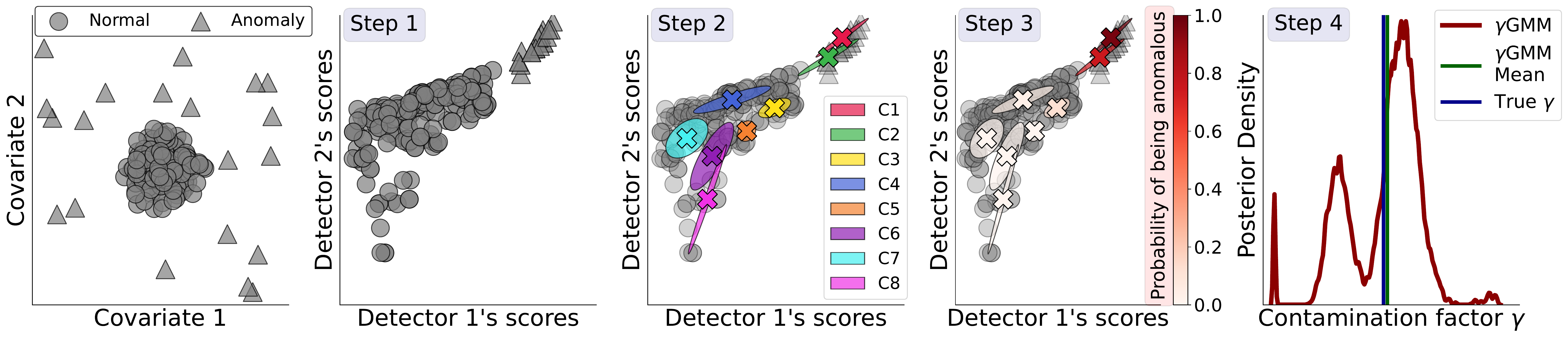}}
\vspace{.1in}
\caption{Illustration of the \ourmethod{}'s four steps on a 2D toy dataset (left plot): we 1) map the 2D dataset into an $M=2$ dimensional anomaly space, 2) fit a DPGMM model on it, 3) compute the components' probability of being anomalous (conditional, in the plot), and 4) derive $\gamma|S$'s posterior. \ourmethod{}'s mean is an accurate point estimate for the true value $\gamma^*$.}\label{fig:foursteps}
\end{figure*}

In this paper, we propose \ourmethod{}, a novel Bayesian approach that estimates the contamination factor's posterior distribution in four steps, which are illustrated in Figure~\ref{fig:foursteps}:
\\
\textbf{Step 1.} Because anomalies may not follow any particular pattern in covariate space, \ourmethod{} maps the covariates $X\in\R^d$ into an $M$ dimensional anomaly space, where the dimensions correspond to the anomaly scores assigned by the $M$ unsupervised anomaly detectors. Within each dimension of such a space, the evident pattern is that ``the higher the more anomalous''. \\
\textbf{Step 2.} We model the data points in the new space $\R^M$ using a Dirichlet Process Gaussian Mixture Model (DPGMM)~\citep{neal1992bayesian,rasmussen1999infinite}.
We assume that each of the (potentially many) mixture components contains either only normals or only anomalies. If we knew which components contained anomalies, we could then easily derive $\gamma$'s posterior as the sum of the mixing proportions $\pi$ of the anomalous components. However, such information is not available in our setting. \\
\textbf{Step 3.} Thus, we order the components in decreasing order, and we estimate the probability of the largest $k$ components being anomalous. This poses three challenges: (a) how to represent each $M$-dimensional component by a single value to sort them from the most to the least anomalous, (b) how to compute the probability that the $k$th component is anomalous given that the $(k-1)$th is such, (c) how to derive the target probability that $k$ components are jointly anomalous. \\
\textbf{Step 4.} \ourmethod{} estimates the contamination factor's posterior by exploiting such a joint probability and the components' mixing proportions posterior.

In the following, we describe these steps in detail.

\subsection{Representing Data Using Anomaly Scores}
\label{subsec:creating_M_dimensional_space}
Learning from an unlabeled anomaly detection dataset has two major challenges. First, anomalies are rare and sparse events, which makes it hard to use common unsupervised methods like clustering~\citep{breunig2000lof}. Second, making assumptions on the unlabeled data is challenging due to the absence of specific patterns in the anomalies, which makes it hard to choose a specific anomaly detector.

Therefore, we use a set of $M$ anomaly detectors to map the $d$-dimensional input space into an $M$-dimensional score space $\R^M$, such that a sample $x$ gets a score $s$: 
\begin{equation*}
    \R^d \ni x \to [f_1(x), f_2(x), \dots, f_M(x)] = s \in \R^M .
\end{equation*}
This has two main effects: (1) it introduces an interpretable space where the evident pattern is that, within each dimension, higher scores are more likely to be anomalous, and (2) it accounts for multiple inductive biases by using multiple arbitrary anomaly detectors.

To make the dimensions comparable, we (independently for each dimension) map the scores $s \in S$ to $\log(s-\min(S)+0.01)$, where the log is used to shorten heavy right tails, and normalize them to have zero mean and unit variance.

\subsection{Modeling the Density with DPGMM}
We use mixture models as basis for quantifying the distribution of the contamination factor, relying on their ability to model the proportions of samples using the mixture weights.
For flexible modeling, we use the DPGMM
\begin{align*}
    s_i &\sim \mathcal{N} (\tilde{\mu}_i, \tilde{\Sigma}_i) \qquad i = 1,\dots, N\\
    (\tilde{\mu}_i, \tilde{\Sigma}_i) &\sim G\\ 
    G &\sim DP(G_0, \alpha)\\
    G_0 &= \mathcal{NIW}(M, \lambda, V, u)
\end{align*}
where $G$ is a random distribution of the mean vectors $\mu_i$ and covariance matrices $\Sigma_i$, drawn from a DP with base distribution $G_0$. We use the explicit representation
$G =  \sum_{k=1}^{\infty} \pi_k \delta_{(\mu_k, \Sigma_k)}(\tilde{\mu}_i, \tilde{\Sigma}_i)$, where $\delta_{(\mu_k, \Sigma_k)}$ is the 
delta distribution at $(\mu_k, \Sigma_k)$ and $\pi_k$ follow the stick-breaking distribution. We set $G_0$ as Normal Inverse Wishart~\citep{nydick2012wishart} with parameters $M, \lambda, V, u$ common to all components. We use variational inference (VI; see e.g. \citet{blei2017variational} for details) for approximating the posterior as VI is computationally efficient and sufficiently accurate for our purposes. Alternative methods (e.g., Markov Chain Monte Carlo \cite{brooks2011handbook}) could also be used but were not considered worth the additional computational effort here.

\paragraph{Choice of DPGMM.} 

DPGMM has two key properties that justify its use over other flexible density models. First, we choose Gaussian distributions over more robust heavy-tailed distributions because isolated samples are likely candidates for outliers, and encouraging the model to represent them using the heavy tails would be counter-productive. Second, the rich-get-richer property of DPs is desirable because we expect some very large components of normals but want to allow arbitrarily small clusters of anomalies. Moreover, the DP formulation allows us to refrain from specifying the number of components $K$. After fitting the model, we only consider the components with at least one observation assigned to them and propagate all the remaining density uniformly over the active components. Thus, for the following steps we can still proceed as if the model was a finite mixture with $\pi$ following a Dirichlet distribution.

\subsection{Estimating the Components' Anomalousness}
We assume that each mixture component either contains only anomalous or only normal samples. All unsupervised methods rely on some assumption on nearby samples sharing latent characteristics, and this cluster assumption is a natural and weak assumption.
If we knew which components contain anomalies, we could directly derive the posterior of the contamination factor $\gamma$ as the sum of the mixing proportions $\pi_k$ of those components. This is naturally not the case, but we need to estimate it in an unsupervised fashion. 

More formally, we estimate the probability that $k$ (out $K$) components are anomalous such that we can derive $\gamma$'s posterior by averaging over all the values $0\le k \le K$. We do this in three steps. Initially, we sort the components of score vectors in decreasing order (by degree of anomalousness), which comes natural from the representation we made in Step $1$ (Sec.~\ref{subsec:creating_M_dimensional_space}). Then, our insight is that the $k$th component can be anomalous only if the $(k-1)$th is such. This points to the estimation of conditional probabilities, i.e., the probability of $c_k =$ ``\textbf{the $k$th component is anomalous}'' given $c_{k-1}$. Finally, the probability that exactly the first $k$ components are anomalous can be obtained using basic rules of probability theory.

\paragraph{Assigning an ordering to the components.}
As initial step for computing the joint probability, we need to design a decreasing ordering map for the components based on their anomalousness.
We do this in a manner that accounts for the uncertainty of the components' parameters to rank high the components that can be reliably identified as anomalous: we want the means to be high but the variance low, to avoid the risk that also samples with low anomaly scores could belong to the component.

We construct the overall ranking using dimension-specific scores because our normalization cannot remove all statistical differences between the different detectors.
Formally, let $r \colon \R^M \times \R^{M\times M} \to \R$ be the function of the mean vector $\mu_k$ and the covariance matrix $\Sigma_k$ that assigns a real value representing the component $k$'s anomalousness. We set $r$ as
\begin{equation}\label{eq:representative_value}
    r\left(\mu^{(z)}_k, \Sigma^{(z)}_k \right) = \frac{1}{M} \sum_{j=1}^M \frac{\mu_k^{j \, (z)}}{1+ \sqrt{\Sigma_k^{j,j \, (z)}}},
\end{equation}
where $\mu_k^{(z)}$ and $\Sigma_k^{(z)}$ are samples from the parameters' posterior distributions of the $k$th component. We obtain a representative value of the whole component by taking the expected value of $r$, i.e. through $\E[r(\mu_k, \Sigma_k)]$. Equation~(\ref{eq:representative_value}) intentionally does not consider inter-dimension correlations, as it remains unclear to us how those should ideally be included and what benefits it would actually provide.

We add $1$ to the component's standard deviation for two reasons. First, if a component contains samples with almost the same covariate values, the standard deviation would be close to $0$ and the ratio would explode towards infinity, masking any effect of the mean. Second, adding $1$ is reasonable because it is equal to the theoretical upper bound of the components' variances, as they are normalized (Sec.~\ref{subsec:creating_M_dimensional_space}).

Without loss of generality, from now on we assume that the components' index $k$ is ordered based on their representative value such that the $k$th component has a higher value (i.e., more anomalous) than the $(k+1)$th component.

\paragraph{Estimating the probability that the $k$th component is anomalous.}
Because the components are sorted by anomalousness, our key insight is that \emph{the $k$th component can be anomalous only if the $(k-1)$th is anomalous}. Formally, 
\begin{equation*}
   \Pp (c_k | \ c_{k-1}) > 0 \ \text{ \& } \ 
   \Pp (c_k | \ \bar{c}_{k-1}) = 0 \quad (1< k \le K)
\end{equation*}
where $\bar{c}_{k-1}$ means ``not $c_{k-1}$''. Moreover, we assume
$\Pp (c_1) \in (0,1)$. That is, we allow for the data to not have anomalies ($<1$) but exclude certain knowledge of no anomalies ($>0$).
This is a sensible assumption because, if one knew for sure that no anomalies are in the data, then we trivially have $\gamma=0$, whereas we still need to allow for the data to be free of anomalies if evidence suggests so.

We estimate the conditional probability as
\begin{equation}\label{eq:sigmoid_function}
    \Pp (c_k | c_{k-1}) = \frac{1}{1+ e^{(\tau + \delta \cdot r(\mu_k, \Sigma_k))}},
\end{equation}
where $\tau$ and $\delta$ are the two hyperparameters of the sigmoid function, which will be carefully discussed in Section~\ref{subsec:posteriordistr}. Note that the principle itself is not restricted to this particular choice of functional form. One could apply any transformation that maps to $[0,1]$, but the detailed derivations of the parameters would naturally be different.

\paragraph{Deriving the components' joint probability.}

Given the conditional probability $\Pp (c_k | \ c_{k-1})$, the joint probability follows from simple steps. Taking inspiration from the sequential ordinal models~\citep{burkner2019ordinal}, our insight is that exactly $k$ components are jointly anomalous if and only if each of them is conditionally anomalous and the $(k+1)$th is not anomalous. We indicate this as $C^* = k$. Essentially, 
\begin{equation}\label{eq:exactProbabilities}
\begin{split}
    \Pp(C^* = k) &\coloneqq \Pp(c_1,\dots,c_k, \bar{c}_{k+1},\dots,\bar{c}_{K}) \\
    &= \Pp(c_1)\prod_{t = 1}^{k-1} \Pp(c_{t+1} | c_t) (1 - \Pp(c_{k+1} | c_k))
\end{split}
\end{equation}
for any $k \le K$, where $\Pp(c_{K+1} \ | c_K) = 0$ by convention.

\subsection{Estimating the Contamination Factor's Distribution}\label{subsec:posteriordistr}
Given the joint probability that the first $k$ components are anomalous (for $k \le K$), the contamination factor $\gamma$'s posterior distribution can be obtained as
\begin{equation}\label{eq:FormulaGammaDistribution}
    p(\gamma | S) \! = \! \sum_{k = 1}^K p(C^* = k) \cdot p\left(\sum_{j = 1}^k \pi_j \bigg| S \right)
\end{equation}
where $p(\sum_{j = 1}^k \pi_j | S)$ is the posterior distribution of the sum of the first $k$ components' mixing proportions, $p(C^* = k)$ are densities WRT the counting measure. Note that $p(\sum_{j = 1}^k \pi_j | S) = \textsc{Beta}(\sum_{j = 1}^k\alpha_j, \sum_{j=k+1}^K \alpha_j )$, if $p(\pi_1 ,\dots,\pi_K| S) = \textsc{Dir}(\alpha_1,\dots,\alpha_K)$~\citep{lin2016dirichlet}.

\paragraph{Setting the sigmoid's hyperparameters $\tau$ and $\delta$.} Introducing new hyperparameters when the task is to estimate the contamination factor $\gamma$'s posterior is risky because setting their value may be as difficult as directly providing a point estimate of $\gamma$. Our key insight is that we can obtain $\tau$ and $\delta$ by asking the user two simple questions: (a) How likely is that no anomalies are in the data? (b) How likely is that a large amount of anomalies occurred, say, more than $t = 15\%$ of the data? Both of these values are supposed to be low. Let's call $p_0$ and $p_{\rm high}$ the two answers. Formally,
\begin{align*}
    p_0 &= 1 - \Pp (c_1) = 1 - \frac{1}{1+ e^{(\tau + \delta \cdot r(\tilde{\mu}_1, \tilde{\Sigma}_1))}}\\
    p_{\rm high} &= \Pp (\gamma \ge t|S ) = \sum_{k = 1}^K \Pp(C^* = k) \cdot \Pp \! \left(\sum_{j = 1}^k \pi_j\ge t|S \right)
\end{align*}
One can use a numerical solver for non-linear equations with linear constraints (e.g., the least square optimizer implemented in \textsc{SkLearn}) to find the values of $\tau$ and $\delta$ that satisfy such constraints. The problem has a unique solution whenever $p_{\rm high} \ge \Pp(\pi_1 \ge t|S)$. 
This holds almost always in our experimental cases, but, in case such a constraint cannot be satisfied, we keep running again the variational inference method (with different starting points) for the DPGMM until the constraint on $p_{\rm high}$ holds. If this cannot happen or does not happen within $100$ iterations, we reject the possibility of too high contamination factors and just set it to $0$.
In the experiments (Q5), we show that changing the $p_0$ and $p_{\rm high}$ does not have a large impact on $\gamma$'s posterior.

\paragraph{Sampling from $\gamma$'s posterior.}
Our estimate of the contamination factor's posterior $p(\gamma | S)$ does not have a simple closed form. However, we can sample from the distribution using a simple process. The DPGMM inference determines an approximation for $p(\pi, \mu, \Sigma | S)$ and all the quantities required for Equations~(\ref{eq:sigmoid_function}),~(\ref{eq:exactProbabilities}),~(\ref{eq:FormulaGammaDistribution}) can be computed based on samples from the approximation. 
Formally, we derive a sample from $p(\gamma|S)$ in four steps by repeating the next operations for all $k \le K$.
First, we draw a sample $\pi_k^{(z)}, \mu_k^{(z)}, \Sigma_k^{(z)}$ from $\pi_k$ (Dirichlet), $\mu_k$ (Normal), $\Sigma_k$ (Inverse Wishart). 
Second, we transform $\pi_k^{(z)}$ by taking the cumulative sum and obtain a sample $\sum_{j = 1}^k \pi_j^{(z)}$. 
Third, we pass $\mu_k^{(z)}$ and $\Sigma_k^{(z)}$ through the sigmoid function ~(\ref{eq:sigmoid_function}) to get the conditional probabilities $\Pp(c_k \ | \ c_{k-1})$, and transform them into the exact joint probabilities $\Pp(C^* = k)$ using the equation~\ref{eq:exactProbabilities}. 
Finally, we multiply the samples following Formula~\ref{eq:FormulaGammaDistribution} and obtain a sample $\gamma^{(z)}$ from $p(\gamma|S)$.


\paragraph{Additional technical details.}
Because our method uses the variational inference approximation, we run it $10$ times and concatenate the samples to reduce the risk of biased distributions due to local minima. Moreover, after sorting the components, we set $\Pp(c_{k} | c_{k-1}) = 0$ for all $k > K' = \argmax \{k \colon \  \E[\sum_{j=1}^k \pi_j] < 0.25 \}$. This has the effect of setting an upper bound of $0.25$ to the contamination factor $\gamma$.
Because anomalies must be rare, we realistically assume that it is not possible to have more than $25\%$ of them. Although ``$0.25$'' could be considered a hyperparameter, this value has virtually no impact on the experimental results. Moreover, note that $\E[\pi_1] \ge 0.25$ cannot occur, as otherwise we could not set the hyperparameters $p_0$ and $p_{\rm high}$.

%% file: texfiles/experiments.tex
\section{Experiments}\label{sec:experiments}
We empirically evaluate two aspects of our method: (a) whether it accurately estimates the contamination factor's posterior, and (b) how thresholding the scores using our method affects the anomaly detectors' performance. To this end, we address the following five experimental questions:
\begin{itemize}
    \item[Q1.] Is the posterior estimate sharp and well-calibrated?
    \item[Q2.] How does \ourmethod{} compare to threshold estimators?
    \item[Q3.] Does a better point estimate of $\gamma$ improve the anomaly detector performance?
    \item[Q4.] What is the impact of the number
    of detectors $M$?
    \item[Q5.] How sensitive the method is to $p_0$ and $p_{\rm high}$?
\end{itemize}


\subsection{Experimental Setup}

\paragraph{Methods.}
We compare the sample mean of \ourmethod{}\footnote{Code and online Supplement are available at: \url{https://github.com/Lorenzo-Perini/GammaGMM}} with $21$ threshold estimators that we cluster into $9$ groups:
\newline
\emph{1. Kernel-based.}
\textsc{Fgd}~\citep{qi2021iterative} and \textsc{Aucp}~\citep{ren2018robust} both use the kernel density estimator to estimate the score density; \textsc{Fgd} exploits the inflection points of the density's first derivative, while \textsc{Aucp} uses the percentage of the total kernel density estimator's AUC to set the threshold;
\newline
\emph{2. Curve-based.}
\textsc{Eb}~\citep{friendly2013elliptical} creates elliptical boundaries by generating pseudo-random eccentricities, while \textsc{Wind}~\citep{jacobson2013robust} is based on the topological winding number with respect to the origin;
\newline
\emph{3. Normality-based.}
\textsc{Zscore}~\citep{bagdonavivcius2020multiple} exploits the Z-scores, \textsc{Dsn}~\citep{amagata2021fast} measures the distance shift from a normal distribution, and \textsc{Chau}~\citep{bol1975chauvenet} follows the Chauvenet’s criterion before using the Z-score;
\newline
\emph{4. Regression-based.}
\textsc{Clf} and \textsc{Regr}~\citep{aggarwal2017introduction} are two regression models that separate the anomalies based on the y-intercept value;
\newline
\emph{5. Filter-based.}
\textsc{Filter}~\citep{hashemi2019filtering}, and \textsc{Hist}~\citep{thanammal2014effective} use the wiener filter and the Otsu's method to filter out the anomalous scores;
\newline
\emph{6. Statistical test-based.}
\textsc{Gesd}~\citep{alrawashdeh2021adjusted}, \textsc{Mcst}~\citep{coin2008testing} and \textsc{Mtt}~\citep{rengasamy2021towards} are based on, respectively, the generalized extreme studentized, the Shapiro-Wilk, and the modified Thompson Tau statistical tests;
\newline
\emph{7. Statistical moment-based.}
\textsc{Boot}~\citep{martin2006evaluation} derives the confidence interval through the two-sided bias-corrected and accelerated bootstrap; \textsc{Karch}~\citep{afsari2011riemannian} and \textsc{Mad}~\citep{archana2015periodicity} are based on means and standard deviations, i.e., the Karcher mean plus one standard deviation, and the mean plus the median absolute deviation over the standard deviation;
\newline
\emph{8. Quantile-based.}
\textsc{Iqr}~\citep{bardet2017new} and \textsc{Qmcd}~\citep{iouchtchenko2019deterministic} set the threshold based on quantiles, i.e., respectively, the third quartile $Q_3$ plus $1.5$ times the inter-quartile region $|Q_3 - Q_1|$, and the quantile of one minus the Quasi-Monte Carlo discreprancy;
\newline
\emph{9. Transformation-based.}
\textsc{Moll}~\citep{keyzer1997using} smooths the scores through the Friedrichs’ mollifier, while \textsc{Yj}~\citep{raymaekers2021transforming} applies the Yeo-Johnson monotonic transformations.

%
%
%
%
%
%
%
%
%

We apply each threshold estimator to the univariate anomaly scores of each detector at a time. 
\emph{We average the contamination factors over the $M$ detectors and use it as the final point estimate for each dataset}.

\paragraph{Data.}
We carry out our study on $20$ commonly used benchmark datasets and additionally $2$ (proprietary) real tasks. The benchmark datasets contain semantically useful anomalies widely used in the literature~\citep{campos2016evaluation}. The datasets vary in size, number of features, and true contamination factor. The online Supplement provides further details.
For the real tasks, our experiments focus on preventing blade icing in wind turbines. We use two public wind turbine datasets, where sensors collect various measurements (e.g., wind speed, power energy, etc.) every $7$ seconds for either $8$ weeks (turbine $15$) or $4$ weeks (turbine $21$). Following~\cite{zhang2018ice}, we construct feature-vectors by taking the average over the time segment of one minute. 

\paragraph{Evaluation metrics.}
We use three evaluation metrics to assess the performance of the methods. 
Contrary to all the threshold estimators, our method estimates the posterior of $\gamma$. Therefore, we measure the \textbf{probabilistic calibration} of \ourmethod{}'s posterior using a QQ-plot with the x-axis representing the expected probabilities and on the y-axis the empirical frequencies. That is, for $v \in [0,0.5]$,
\begin{align*}
    \text{Expected Prob.} &=  \Pp\left(\gamma^* \in [q(0.5-v), q(0.5+v)]\right) = 2v\\
    \text{Empirical Freq.} &= \frac{\left|\left\{\gamma \in [q(0.5-v), q(0.5+v)]\right\}\right|}{\# \text{experiments}},
\end{align*}
where $q(u)$ is the quantile at the value $u$ of our distribution, for $u \in [0,1]$, and $\gamma^*$ refers to the true dataset's contamination factor. 
For evaluating the point estimate of the methods, we use the \textbf{mean absolute error} (MAE) between the method's point estimate and the true value. Finally, we measure the impact of thresholding the scores using the methods' point estimate through the $F_1$ score~\citep{goutte2005probabilistic}, as common metrics like the Area Under the ROC curve and the Average Precision are not affected by different thresholds.
Specifically, for $m = 1,\dots, M$, we measure the \textbf{relative deterioration of the $F_1$ score}:
\begin{equation*}
    F_1 \text{ deterioration} = \frac{F_1 (f_m, D,\gamma^*) - F_1 (f_m, D,\hat{\gamma})}{F_1 (f_m, D,\hat{\gamma})}
\end{equation*}
where we compute the $F_1$ score on the dataset $D$ using the anomaly detector $f_m$, and either the true value $\gamma^*$ or an estimate $\hat{\gamma}$ to threshold the scores. The $F_1$ deterioration of a method is (mostly) negative, and the higher the better.

\paragraph{Setup.}
In the experiments we assume a transductive setting~\citep{campos2016evaluation,scott2008transductive,toron2022transductgan}, where a dataset $D$ is used both for training and testing. This is the typical setting of anomaly detection~\citep{breunig2000lof,scholkopf2001estimating,angiulli2002fast,liu2012isolation} because the absence of labels and patterns (for the anomaly class) avoids overfitting issues.

For each dataset, we proceed as follows: 
(i) use a set of $M$ anomaly detectors to assign the anomaly scores $S$ to each observation in the dataset $D$; 
(ii) map each anomaly score $s \in S$ to $\log(s-\min(S)+0.01)$ and normalize them to have mean equal to $0$ and standard deviation equal to $1$; 
(iii) either use our method to estimate the contamination factor's posterior and extract the posterior mean as point estimate $\hat{\gamma}$, or use one of the threshold estimators to directly obtain a point estimate $\hat{\gamma}$ of the contamination factor (see methods paragraph above); 
(iv) evaluate the point estimates using the mean absolute error (MAE) between such estimate and the true value $\gamma^*$;
(v) use the contamination factor's point estimate to threshold the anomaly scores of each of the $M$ anomaly detectors $f_m$ (individually); 
(vi) finally, we measure the $F_1$ score and compute the relative deterioration.

\paragraph{Hyperparameters, anomaly detectors and priors.}

Our method introduces two new hyperparameters: $p_0$ and $p_{\rm high}$. We both of them set to $0.01$ as default value because extremely high contamination, as well as no anomalies, are unlikely events. We will experimentally check the impact of these two hyperparameters in Q5.

We use $10$ anomaly detectors with different inductive biases~\citep{soenen2021effect}: 
\textsc{kNN}~\citep{angiulli2002fast} assumes that the anomalies are far away from normals,
\textsc{IForest}~\citep{liu2012isolation} assumes that the anomalies are easier to isolate, 
\textsc{LOF}~\citep{breunig2000lof} exploits the examples' density, 
\textsc{OCSVM}~\citep{green2001modelling} encapsulates the data into a multi-dimensional hypersphere, 
\textsc{Ae}~\citep{chen2018autoencoder} and \textsc{VAE}~\citep{kingma2013auto} use the reconstruction error as anomaly score function in a, respectively, deterministic and probabilistic perspective, 
\textsc{LSCP}~\citep{zhao2019lscp} is an ensemble method that selects competent detectors locally, 
\textsc{HBOS}~\citep{goldstein2012histogram} calculates the degree of anomalousness by building histograms, 
\textsc{LODA}~\citep{pevny2016loda} is an ensemble of weak detectors that build histograms on randomly generated projected spaces, and 
\textsc{COPOD}~\citep{li2020copod} is a copula based method. All these methods are implemented in the python library PyOD~\citep{zhao2019pyod}.

The threshold estimators are implemented in \textsc{PyThresh}\footnote{Link: \url{https://github.com/KulikDM/pythresh}.} with default hyperparameters.
Finally, the \textsc{DPGMM} is implemented in \textsc{sklearn}: we use the Stick-breking representation~\citep{dunson2008kernel}, with $100$ as upper bound of $K$. We set the means' prior to $0$, and the covariance matrices' prior to identities of appropriate dimension. We opt for such (in our context) weakly-informative priors because sensible prior knowledge of the \textsc{DPGMM} hyperparameters is hard to come by in practice.

\subsection{Experimental Results}
\begin{figure*}[h]
\vspace{.1in}
\centerline{\includegraphics[width=1\textwidth]{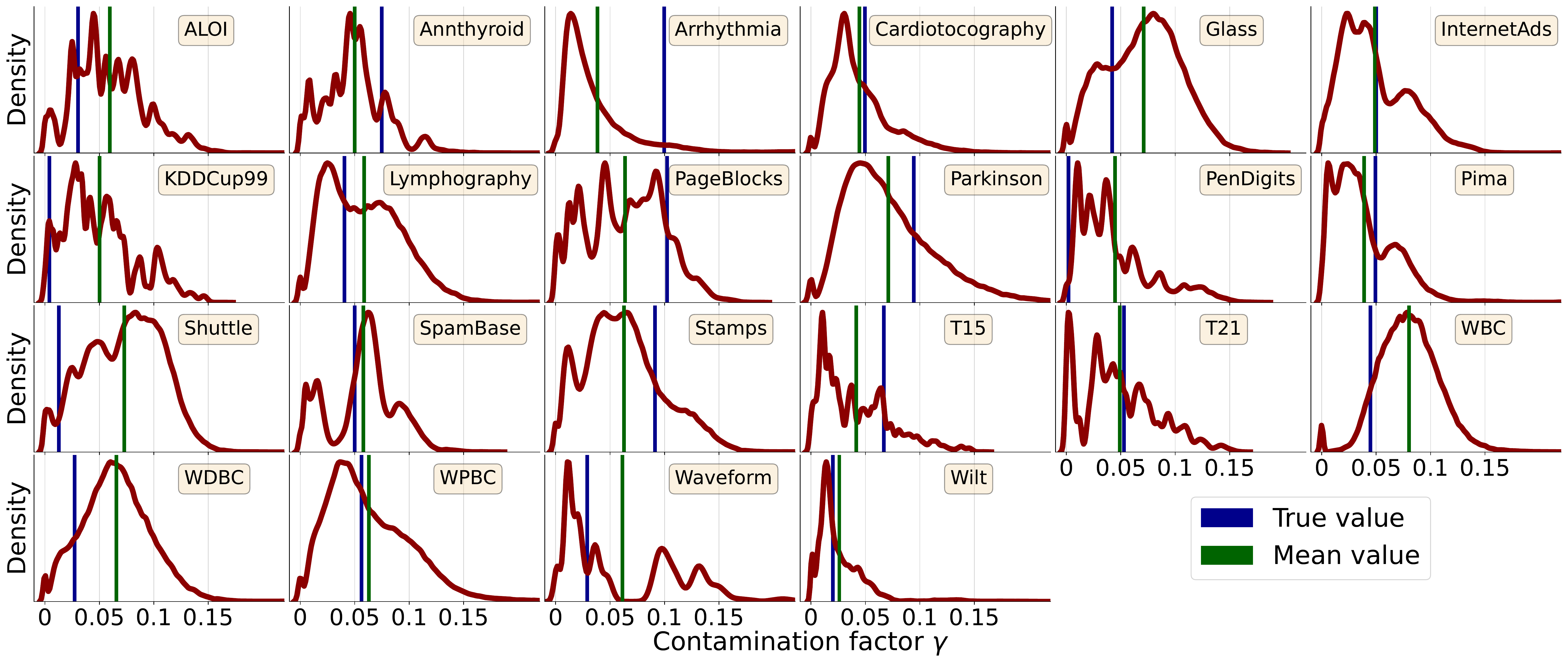}}
\vspace{.1in}
\caption{Illustration of how \ourmethod{} estimates $\gamma$'s posterior distribution (red) on all the $22$ datasets. The blue vertical line indicates the true contamination factor, while the green line is the posterior's mean.
}\label{fig:10densities}
\end{figure*}


\begin{figure}[h]
\vspace{.1in}
\centerline{\includegraphics[width=0.4\textwidth]{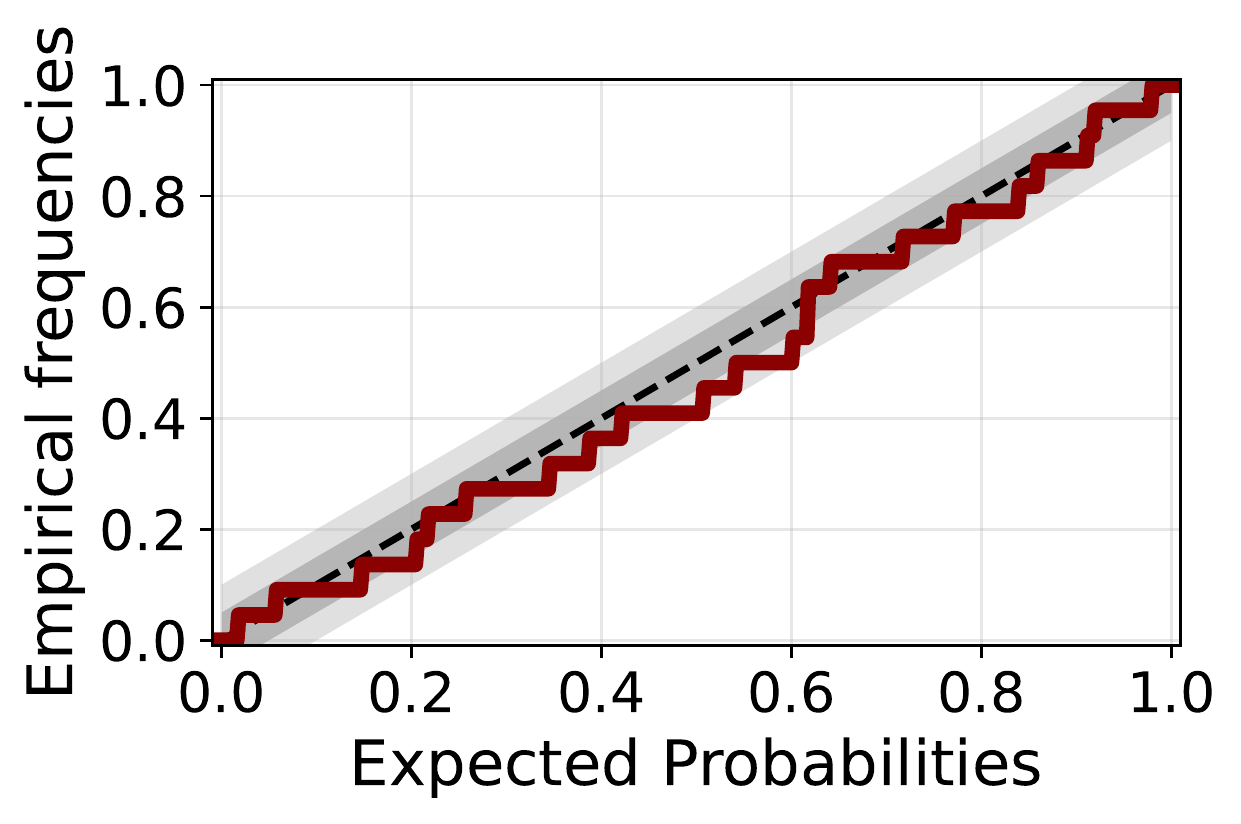}}
\vspace{.1in}
\caption{QQ-plot of \ourmethod{}'s distribution estimate. The black dashed line illustrates the perfect calibration, while shades indicate a deviation of $5\%$ (dark) and $10\%$ (light) from the black line.}\label{fig:qqplot_our_distribution}
\end{figure}

\paragraph{Q1. Does our method estimate a sharp and well-calibrated posterior of $\gamma$?} Figure~\ref{fig:10densities} shows the contamination factor $\gamma$'s posterior estimated by our method on the $22$ datasets. In several cases (e.g., WPBC, Cardio, SpamBase, Wilt and T21), the distribution looks accurate as $\gamma$'s true value (blue line) is close to the posterior mean (i.e., the expected value, the green line). On the contrary, some datasets (e.g., Arrhythmia, Shuttle, KDDCup99, Parkinson, Glass) obtain less accurate distributions: although $\gamma$'s true value sometimes falls on low-density regions (Arrhythmia, Shuttle), in many cases it would be quite likely to sample the true value from our posterior (KDDCup99, Parkinson, Glass), which makes the density still quite reliable.

Figure~\ref{fig:qqplot_our_distribution} shows the calibration plot. 
The posterior is well-calibrated as it is very close to the dashed black line indicating a perfectly calibrated distribution. The empirical frequencies deviate from the real probabilities by less than $5\%$ (dark shadow grey) in more than $76\%$ of the cases, while never deviating by more than $10\%$ (light shadow grey).

\paragraph{Q2. How does \ourmethod{} compare to the threshold estimators?}
\begin{figure}[h]
\vspace{.1in}
\centerline{\includegraphics[width=0.5\textwidth]{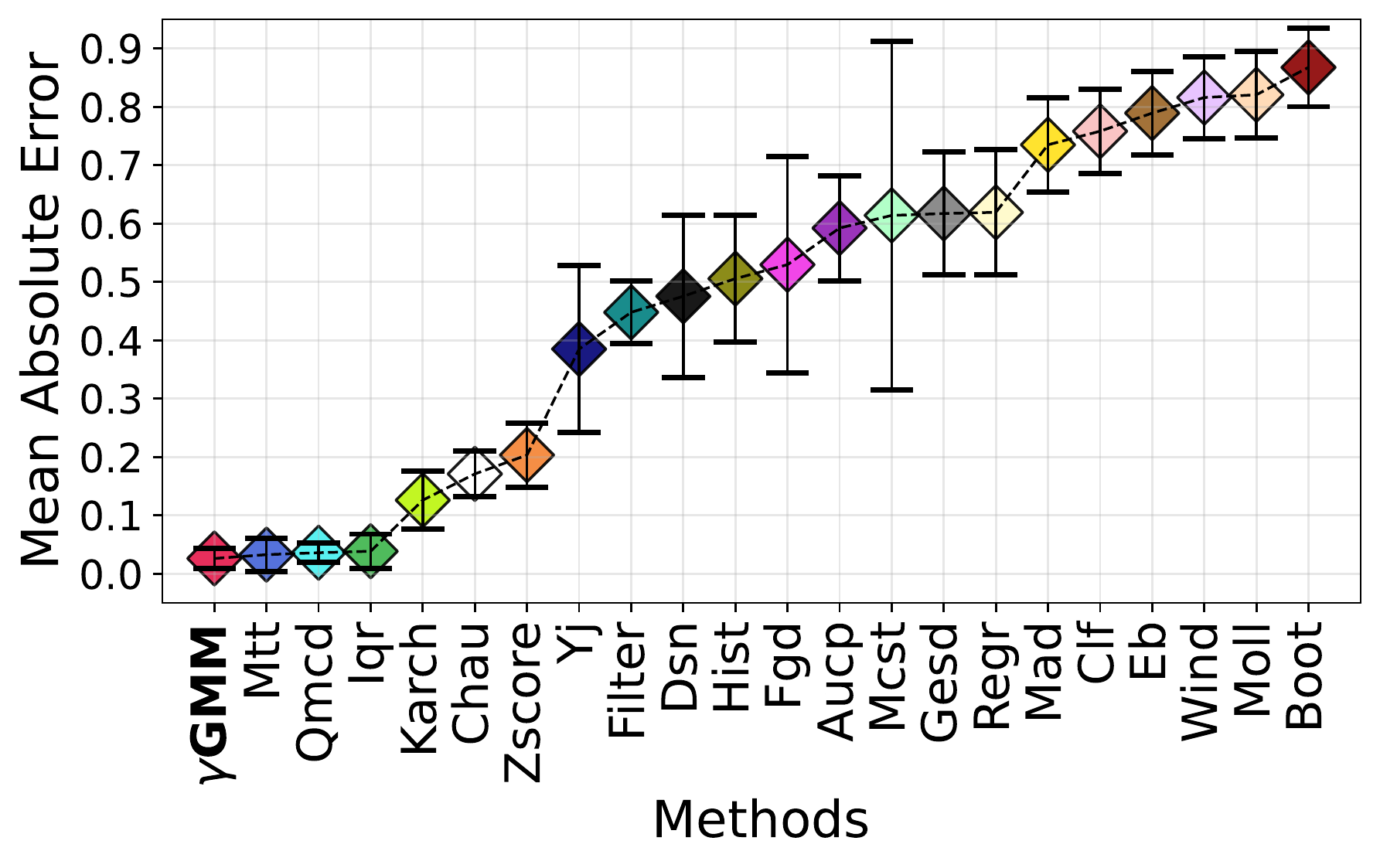}}
\vspace{.1in}
\caption{Average MAE ($\pm$ std.) of \ourmethod{}'s sample mean compared to the other methods. Our method has the lowest (better) average, which is $20\%$ lower than the runner-up.}\label{fig:MAE_baselines}
\end{figure}

We take \ourmethod{}'s posterior mean as our best point estimate of $\gamma$ and compare such value to the point estimates obtained from the threshold estimators. Figure~\ref{fig:MAE_baselines} illustrates the ordered MAE (mean $\pm$ std.) between the methods' estimate and the true $\gamma$. On average, \ourmethod{} obtains a MAE of $0.026$ that is $20\%$ lower than the best runner-up \textsc{Mtt} and $27\%$ lower than the third best method \textsc{Qmcd} (MAE of $0.033$ and $0.036$). For each experiment, we rank the methods from the best (position $1$, lowest MAE) to the worst (position 22, greatest MAE). Our method has the best average rank ($2.13 \pm 1.04$).
Moreover, \ourmethod{} ranks first $8$ times ($\approx 36\%$ of the cases), and for $13$ times ($\approx 60\%$ of the cases) it is in the top two. The next best method, \textsc{Mtt}, ranks first in $6$ cases with an average rank of $2.30 \pm 1.10$.

\paragraph{Q3. Does a better contamination improve the anomaly detectors' performance?}
\begin{figure}[h]
\vspace{.1in}
\centerline{\includegraphics[width=0.5\textwidth]{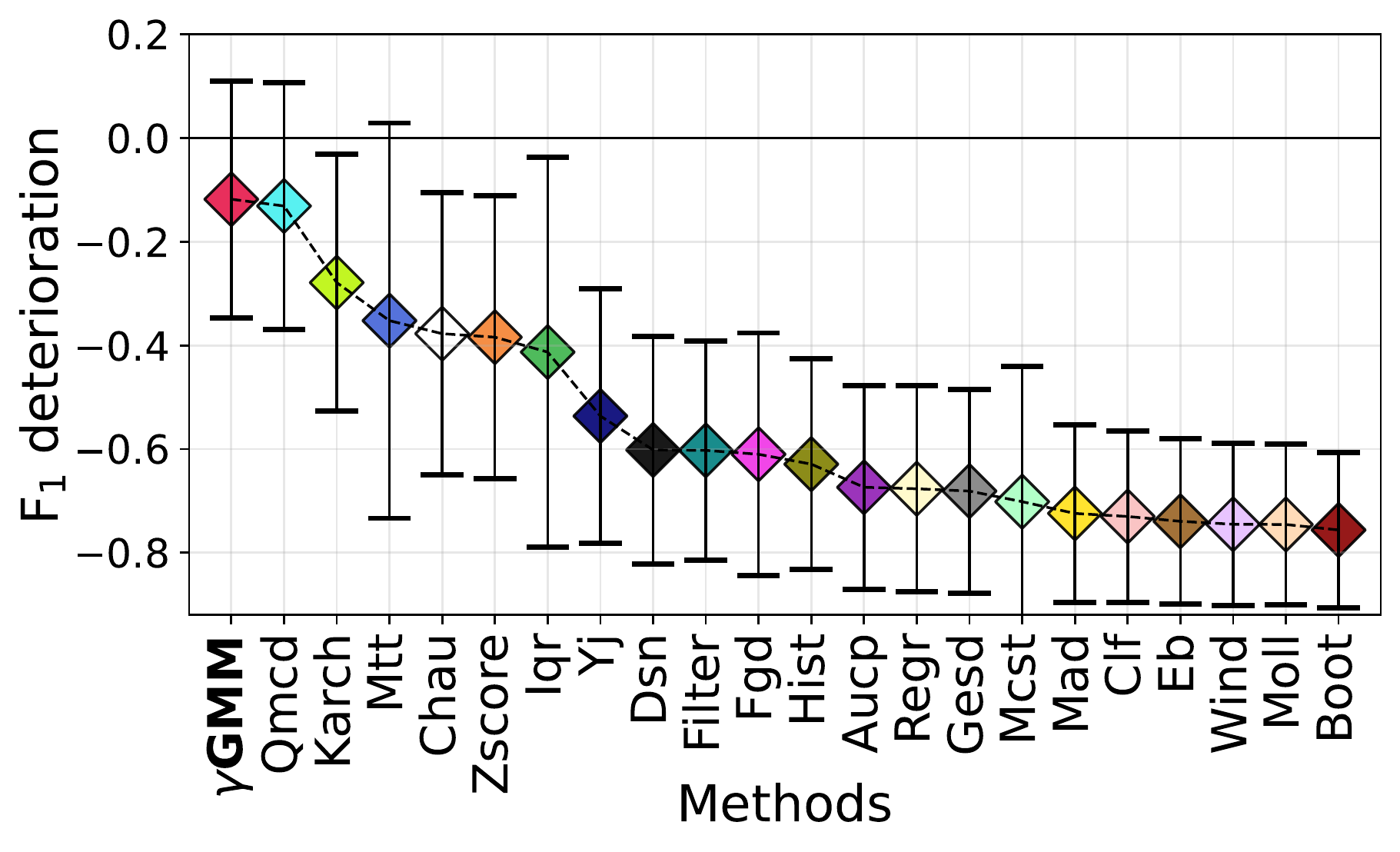}}
\vspace{.1in}
\caption{ $F_1$ deterioration (mean $\pm$ std) for each method, where the higher the better. \ourmethod{} ranks as best method, obtaining $\approx 10\%$ higher average than \textsc{Qmcd}.}\label{fig:F1_baselines}
\end{figure}

We use \ourmethod{}'s posterior mean as a point estimate to measure the $F_1$ score of the anomaly detectors because sampling from the distribution would not imply a fair comparison against the other methods that can only provide a point estimate. Moreover, anomaly detectors that fail to rank the samples accurately perform poorly even when using the correct $\gamma$. Since our focus is studying the effect of $\gamma$, for each dataset $D$, we compare $F_1$ scores only over the detectors that achieve the greatest $F_1$ score using the true contamination factor $\gamma^*$, i.e. $\argmax_{f_m} \left\{F_1 (f_m, D, \gamma^*)\right\}$. The online Supplement contains the list of detectors used for each experiment.

Figure~\ref{fig:F1_baselines} shows the average ($\pm$ std.) deterioration for each of the methods. On average, \ourmethod{} has the best $F_1$ deterioration ($-0.117 \pm 0.228$) that is around $10\%$ better than the runner-up \textsc{Qmcd} ($-0.131 \pm 0.238$), and $58\%$ better than the next best \textsc{Karch} ($-0.279 \pm 0.248$). 
For $25\%$ of the cases we get higher
$F_1$ score with \ourmethod{} than when using the true $\gamma^*$. This is due to the (still incorrect) ranks made by the detectors, which achieve better performance with slightly incorrect contamination factors. The online Supplement provides further details on how the methods perform in terms of false alarms and false negatives.

\paragraph{Q4. What is the impact of $M$ on $\gamma$'s posterior?} 
\begin{figure}[h]
\vspace{.1in}
\centerline{\includegraphics[width=0.4\textwidth]{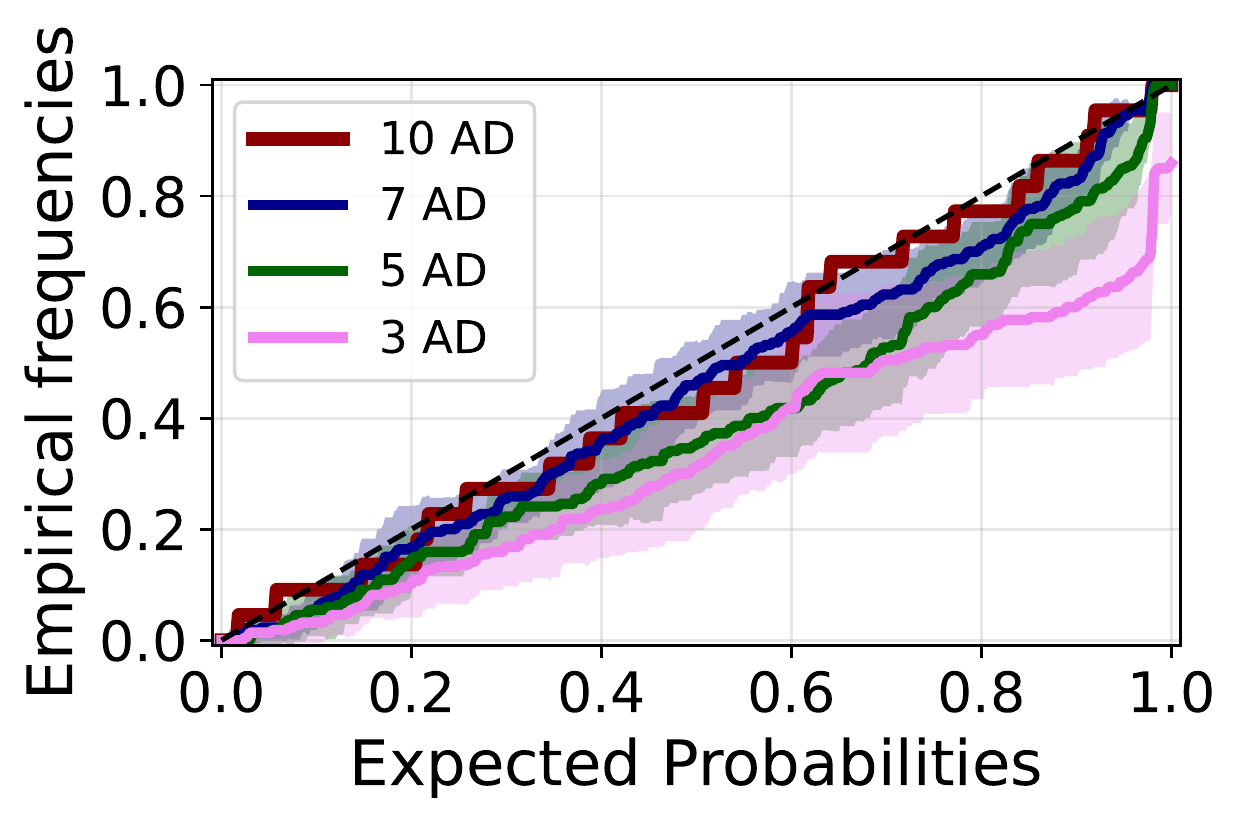}}
\vspace{.1in}
\caption{QQ-plot comparing the calibration curves of \ourmethod{} when a different number $M$ of detectors is used. The colored shades report the uncertainty obtained by randomly sampling the detectors from a set of $10$ detectors. The plot shows that the higher the number of detectors, the more calibrated the distribution.}\label{fig:varyingdetectors}
\end{figure}

In the previous experiments, we used $M=10$ detectors. We evaluate the effect of $M$ by running all the experiments $10$ times with (different) randomly chosen detectors for $M = 3, 5, 7$.
Figure~\ref{fig:varyingdetectors} shows that the calibration suffers if using fewer detectors, but already $M=5$ let the method work fairly well. The variance of the results (over repeated experiments) also increases for lower $M$.

\paragraph{Q5. Impact of the hyperparameters $p_0$ and $p_{\rm high}$.}
\begin{figure}[h]
\vspace{.1in}
\centerline{\includegraphics[width=0.5\textwidth]{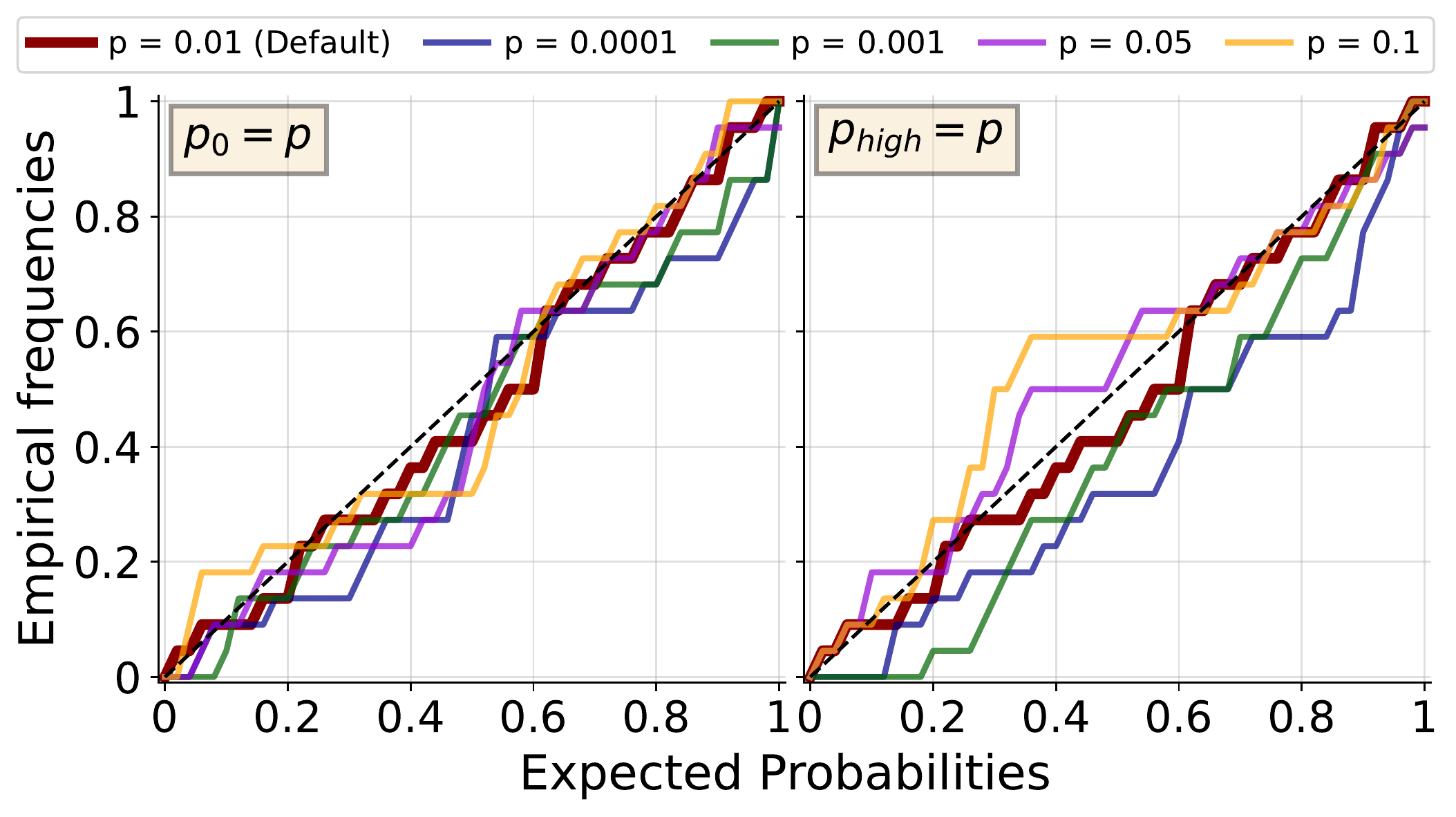}}
\vspace{.1in}
\caption{QQ-plot showing how calibrated \ourmethod{}'s posterior mean would be if we varied $p_0$ (left) and $p_{\rm high}$ (right). While $p_0$ does not have a large impact on the method, the empirical frequencies slightly under (over) estimate the expected probabilities for low (high) values of $p_{\rm high}$.} \label{fig:varying_p0phigh}
\end{figure}

We evaluate the impact of $p_0$ and $p_{\rm high}$ by running the experiments with smaller and larger values than $0.01$: we vary, one at a time, $p_0$, $p_{\rm high} \in [0.0001, 0.001, 0.05, 0.1]$ and keep the other set as default. Figure~\ref{fig:varying_p0phigh} shows the QQ-plot for $p_0$ (left) and $p_{\rm high}$ (right). In both cases, smaller hyperparameters lead to slightly under-estimated expected probabilities. Overall, our method is robust to different values of $p_0$, while $p_{\rm high}$ affects the calibration slightly more. Comparing the resulting $8$ variants of \ourmethod{} in terms of MAE, we conclude that the posterior means produce similar values to our default setting, obtaining an MAE that varies from $0.252$ ($p_{\rm high} = 0.001$, the best) to $0.32$ ($p_{0} = 0.0001$, the worst).

%% file: texfiles/conclusion.tex
\section{Conclusion}\label{sec:conclusion}

The literature on anomaly detection has focused on unsupervised algorithms, but largely ignored practical challenges in their application. The algorithms are evaluated on performance metrics focusing on the ranking of the samples (e.g., AUC), and the ultimate choice of detecting the actual anomalies by thresholding the predictions is left to the practitioners. They lack good means for thresholding and thus often resort to using labels for such goal. This largely defeats the point of using unsupervised methods.

We presented the first practical method for estimating the posterior distribution of the contamination factor $\gamma$ in a completely unsupervised manner. We empirically demonstrated on $22$ datasets that our mean estimates effectively solve the question of where to threshold the predictions. We outperform all $21$ comparison methods and show that the gap in detection accuracy between our estimate and the ground truth (available for these benchmark datasets) is small.

Besides solving the practical question of thresholding the predictions, we seek to persuade the anomaly detection community of the usefulness of a fully probabilistic solution for the problem. Especially in unsupervised settings, it would be completely unreasonable to expect the contamination factor could be identified exactly, but rather we need to characterize its uncertainty. However, we are not aware of any previous works even attempting this. As shown in Fig.~\ref{fig:10densities}, the posterior distribution of $\gamma$ may not only be wide but also multi-modal. Communicating these aspects to the practitioner is critical so that they can e.g. use additional domain knowledge to interpret the alternatives. We showed that our estimates have near-perfect calibration over the broad range of datasets and hence can be relied on in practical use. 

On first impression, the success of our method in solving this challenging and seemingly ill-posed problem may seem surprising. 
However, it can be attributed to a careful choice of strong inductive biases built into the underlying probabilistic model. We argue that all of the following elements are necessary, each substantially contributing to the overall success: (i) representing the data in the space of anomaly detector scores defines a meaning for the dimensions and allows borrowing inductive biases of arbitrary detector algorithms, (ii) the mixture model encodes a natural clustering assumption for both the normal samples and the anomalies, (iii) the ordering used for determining the final distribution incorporates both the location and shape of the mixture components in a carefully balanced manner, and (iv) the transformation from the ordering to probabilities is robustly parameterized via just two intuitive hyperparameters, enabling use of the same defaults for all cases.

%% file: texfiles/acks.tex
\section*{Acknowledgments}
This work was done during LP's research visit to the University of Helsinki, funded by the Gustave Bo\"{e}l - Sofina Fellowship (grant V407821N). Moreover, this work is supported by (LP) the FWO-Vlaanderen aspirant grant 1166222N and by the Flemish government under the “Onderzoeksprogramma Artificiële Intelligentie (AI) Vlaanderen” program, (PB) the Deutsche Forschungsgemeinschaft (DFG, German Research Foundation) under Germany’s Excellence Strategy - EXC 2075 – 390740016, (AK) the Academy of Finland (grants 313125 and 336019), the Flagship program Finnish Center for Artificial Intelligence (FCAI), and the Finnish-American Research and Innovation Accelerator (FARIA).